\documentclass[runningheads]{llncs}
\usepackage{cite}
\usepackage{amsmath,amssymb,amsfonts}
\usepackage{algorithmic}
\usepackage{graphicx}
\usepackage{lineno}
\usepackage{xcolor}

\usepackage{subcaption}
\usepackage{algorithm}
\usepackage{url}
\usepackage{balance}
\usepackage{stfloats}
\usepackage{multirow}
\usepackage[flushleft]{threeparttable}

\newcommand{\Methodname}{Complementary Ensemble Learning }
\newcommand{\argmax}{{\operatorname{arg}\,\operatorname{max}}\;}
\newcommand{\x}{$\times$}
\newcommand{\multirot}[3]{\multirow{#1}{*}{\rotatebox{#2}{ #3 } }}  

\begin{document}
\title{ \Methodname \\}

%

\author{ Hung Nguyen \inst{1}
		\and J. Morris Chang \inst{2}}
	
\institute{Department of Electrical Engineering,
		University of South Florida, \\Tampa, Florida, USA \inst{1,2}
\\	
\email{nsh@usf.edu \inst{1}}\\ 
\email{chang5@usf.edu \inst{2}} 
}

\maketitle

\begin{abstract}
To achieve high performance of a machine learning (ML) task, a deep learning-based model must implicitly capture the entire distribution from data. Thus, it requires a huge amount of training samples, and data are expected to fully present the real distribution, especially for high dimensional data, e.g., images, videos. In practice, however, data are usually collected with a diversity of styles, and several of them have insufficient number of representatives. This might lead to uncertainty in models' prediction, and significantly reduce ML task performance.

In this paper, we provide a comprehensive study on this problem by looking at model uncertainty. From this, we derive a simple but efficient technique to improve performance of state-of-the-art deep learning models. Specifically, we train auxiliary models which are able to complement state-of-the-art model uncertainty. As a result, by assembling these models, we can significantly improve the ML task performance for types of data mentioned earlier. While slightly improving ML classification accuracy on benchmark datasets (e.g., 0.2\% on MNIST), our proposed method significantly improves on limited data (i.e., 1.3\% on Eardrum and 3.5\% on ChestXray).  

\keywords{deep learning, ensemble learning, limited data, data diversity }
\end{abstract}

\section{Introduction}
The explosion of big data in high dimension urged machine learning technologies to develop dramatically, especially deep learning-based (DL) techniques (e.g., convolutional neural network, recurrent neural network). Several DL models have been proposed to tackle machine learning tasks, and they have proved the capability to work on benchmark datasets containing a huge amount of data (e.g., Imagenet, MNIST, Cifar100). In practice, however, the diversity of data has never ended. For example, in a handwritten digit image classification problem, the digit images (even belonging to the same class) still have a variety of styles and shapes. For capturing all features of a class distribution, data need to be well presented for each style and also cover all styles within the class. However, it is likely infeasible to collect such well-presented data in practice. In fact, datasets are usually dominated by popular styles due to the density of those styles' representatives contained in the training datasets. This might cause some levels of uncertainty for a particular model and reduce its performance. In this study, we focus on studying such type of diverse image data (we use the term data in the following sections to imply this type of data) and propose a novel technique of improvement for state-of-the-art deep learning models.    

There is a number of works \cite{8285168,NIPS7278,tanghe_2015,barua_2014} aiming to tackle the problem of lacking diverse data, in which they try to balance the data among classes such as sampling more data or generating synthetic data to reach an equilibrium between classes. These techniques, however, focus on the unbalanced data problem between classes, they have not considered the data within a class. There is also effort to tackle the data diversity problem within class by extracting important features such as dimension reduction \cite{nguyen_2020,6844831}, feature extraction, denoising auto-encoder \cite{denoise,li_2015}, etc. These methods do show some improvement on ML task performance, but these are dominated by the majority of training data as most of information is extracted from the majority. For data that have limited representatives in different styles, the improvement is insignificant.   
  
From the analysis which will be discussed in Section \ref{sec:analysis}, our observation reveals that a portion of model uncertainty caused by data styles barely appear in the training data. This is because learned models' parameters might be dominated by the most popular styles in training data. Consequently, this resists a model from learning data diversity. To tackle this problem, one can use expert knowledge to carefully design a model for these specific datasets by considering different data styles or sampling more samples to balance the data. These approaches, however, are costly, and in some cases, it is just infeasible. In this study, we alleviate this problem by simply complementing the weakness of state-of-the-art models, in which we generate complementary models for specific data area in high dimension space that the others might be not confident. 

Our work has two main contributions:
\begin{itemize}
	\item Exploration of the relationship between data diversity and convolutional neural network uncertainty. 
	\item A proposed technique, namely Complementary Ensemble Learning aiming to alleviate the problem of data diversity. 
\end{itemize}

The following sections in this paper are organized as follows. Section \ref{sec:preliminaries} reviews knowledge about heatmap and our definition of model confidence. Section \ref{sec:analysis} is our analysis on model uncertainty and data diversity. In Section \ref{sec:method}, we proposed a technique named \Methodname to alleviate the problem of data diversity. The experiment results are presented in Section \ref{sec:conclusion} and is followed by our conclusion.
 
\section{Preliminaries}
\label{sec:preliminaries}
In this study, we evaluate model's certainty in a multi-class classification task by a confidence score parameter which will be defined in the following section. We also use Heatmap as a visual method to explore how a model look at its input and what could cause model uncertainty.  
      
\subsection{Confidence Score}
We use a term called Confidence Score to present how confident a model prediction is. We assume models use softmax for output layers and result in a probability-like prediction ranged from 0 to 1 for each class. Let $[p_1^i,..p_{n-1}^j,p_{n}^k]$ be the sorted probability-like prediction for $n$ classes. $i$, $j$ and $k$ indicate classes corresponding to the prediction values of $p_1$, $p_{n-1}, p_{n}$  and As a result, $p_{n}$ is the largest value, and the final predicted class is class $k$. A Confidence Score (CS) is defined as the difference between the largest value and the second largest value. The confidence score can be computed as $CS =  p_{n}- p_{n-1}$.      

\subsection{Heatmap}
In \cite{heatmap}, heatmap was introduced as an effort to visually understand convolutional network properties. The method attempts to perform an invert function through hidden layers and reach back to the input layer. Thus, it results in a map which has similar size to the input image. The map presents the most attentive area where the model focuses on the image. In this study, we implement the method to understand where is the most important parts in the input image for each class and how it is applied to different styles of images. 

In this section, we will explore model's confidence and how data diversity in a training set could negatively affect the model performance. For better visualization, our experiments in this section will be conducted on MNIST dataset which includes 60,000 handwritten digit images from 0-9. It is split into training part (50,000 samples) and validation part (10,000 samples). We then build a basic Convolutional Neural Network (CNN) following the MNIST example in Keras website \cite{mnist_keras}, which uses three convolutional layers, one dense hidden layer, and two dropout layers. All hidden layers are activated by Relu functions, and Softmax is used in the output layer.

\section{Model behaviors on data diversity} \label{sec:analysis}
\begin{figure}[]
	\begin{subfigure}[]{0.5\textwidth}
		\includegraphics[width=\linewidth, trim=100 50 100 100,clip]{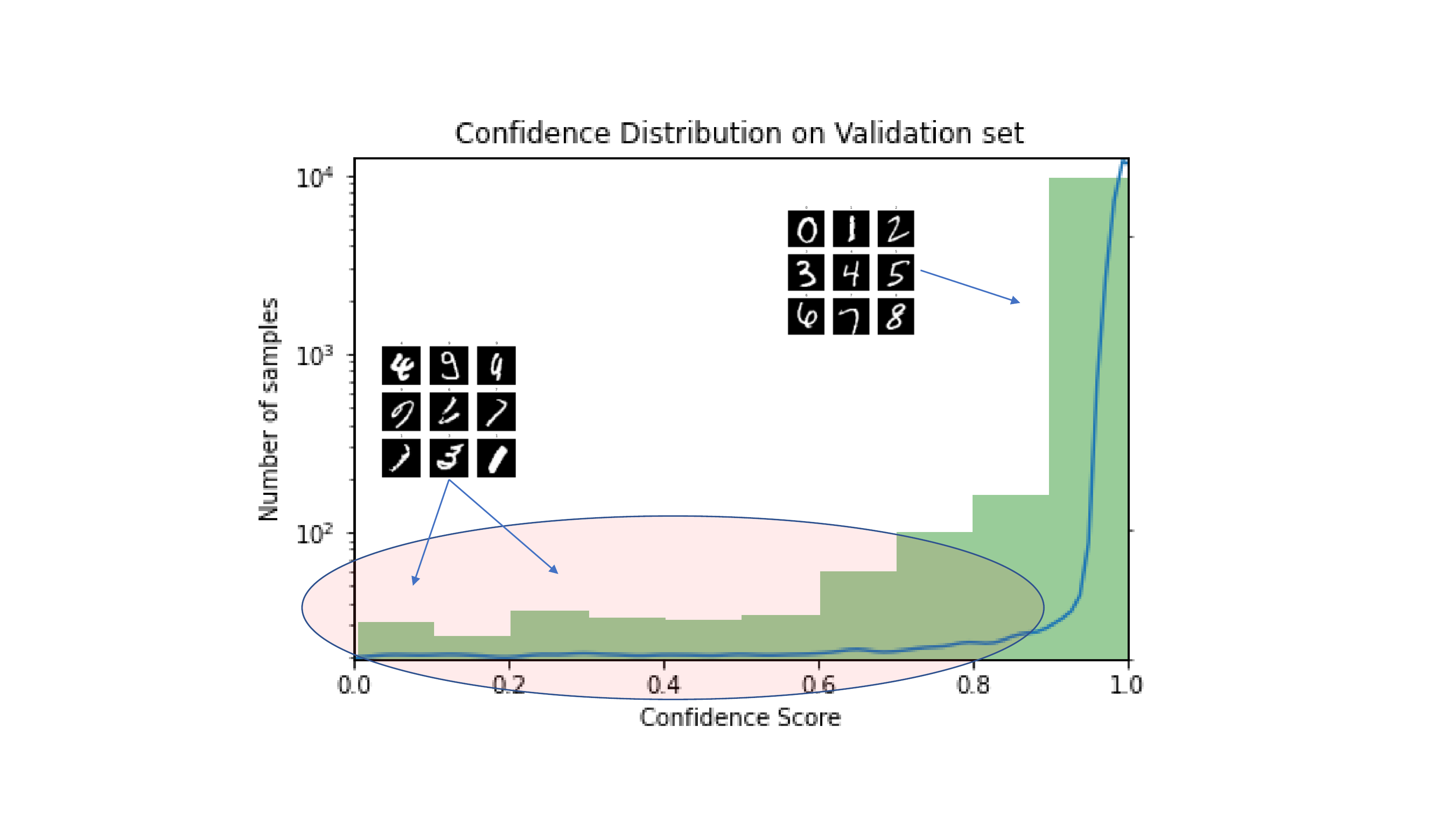}
		\caption{Model Confidence Score Histogram}
		\label{fig:confidence}
	\end{subfigure}
	\begin{subfigure}[]{0.5\textwidth}
		\includegraphics[width=\linewidth, trim=220 120 220 120,clip]{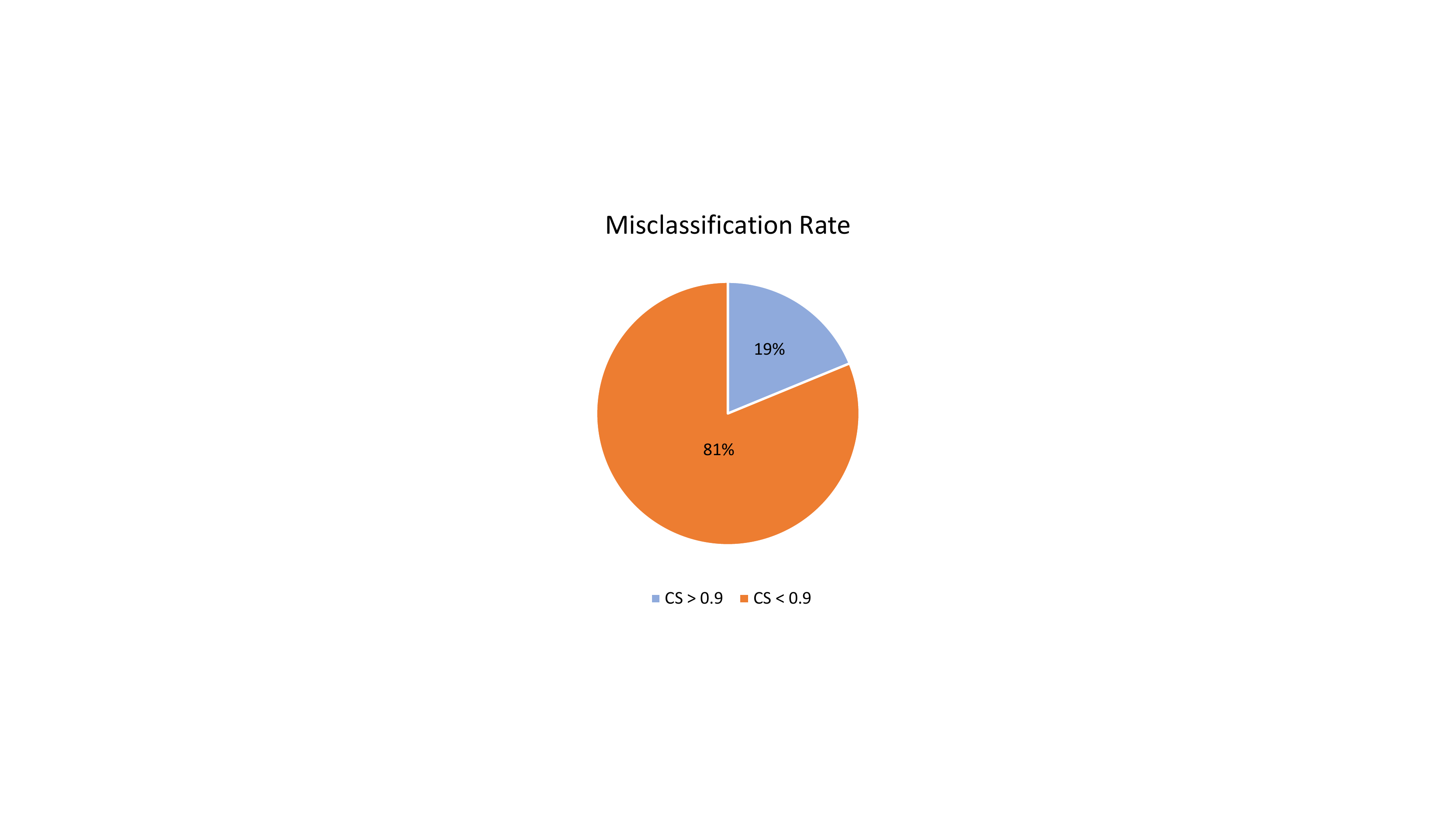}
		\caption{Misclassification rate by Confidence Score }
		\label{fig:misclassification}
	\end{subfigure}
	\caption{Model result}
\end{figure}

\subsection{Model Uncertainty}\label{sec:uncertainty}
This study prompts that the diversity of styles in training data might cause model uncertainty and lead to model misclassification. We are modeling the uncertainty of a deep learning classifier based on confidence score (CS). Figure \ref{fig:confidence} shows the histogram of model prediction on validation set (10,000 samples) in terms of confidence score (CS).

The result demonstrates that most of the samples are predicted with a high confidence score. More specifically, 9489 samples are predicted with CS $\geq$ 0.9, and 511 samples are predicted as CS $<$ 0.9. On the other hand, Figure \ref{fig:misclassification} illustrates model misclassification proportion with respect to confidence score, and it indicates that 81\% of misclassification are classified with low confidence scores. The results of the experiment found clear support for our assumption that model misclassificaion mainly occurs in the model uncertainty area. Thus, this prompts us to pay more attention on the model uncertainty and theoretical reasons causing it, which will be analyzed in \ref{sec:heatmap}.   

\subsection{Model Attention Map}\label{sec:heatmap}

\begin{figure}[H]
	\begin{subfigure}[]{0.5\linewidth}	
		\includegraphics[width=\linewidth, trim=70 210 50 200,clip]{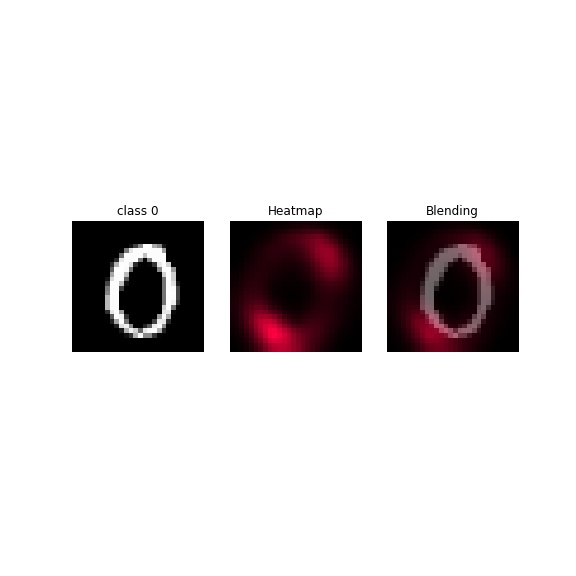}\\	
		\includegraphics[width=\linewidth, trim=70 210 50 200,clip]{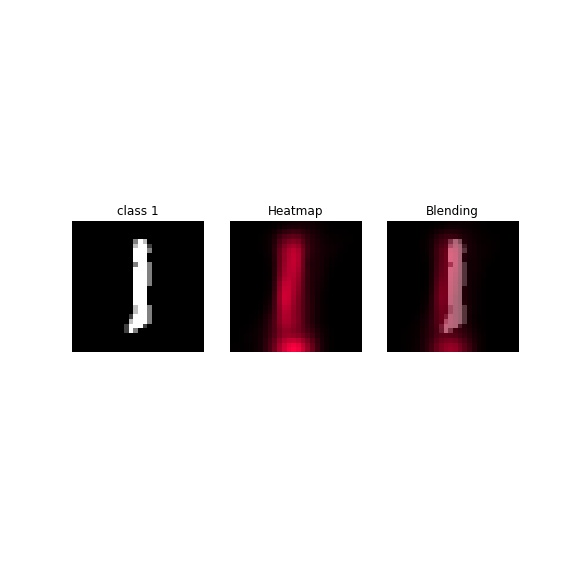}\\
		\includegraphics[width=\linewidth, trim=70 210 50 200,clip]{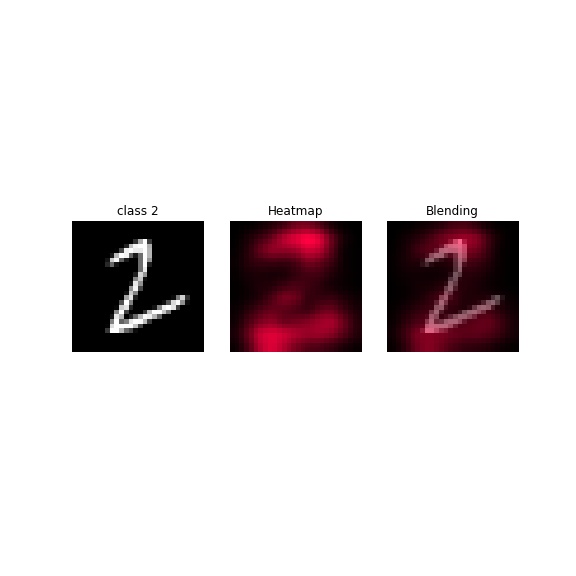}\\	
		\includegraphics[width=\linewidth, trim=70 210 50 200,clip]{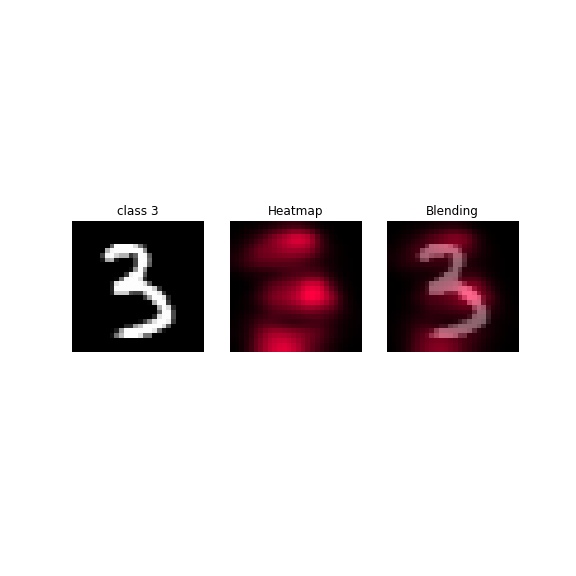}	
		\caption{High confidence samples}
		\label{fig:heatmap1}
	\end{subfigure}	
	\begin{subfigure}[]{0.5\linewidth}
		\includegraphics[width=\linewidth, trim=70 210 50 200,clip]{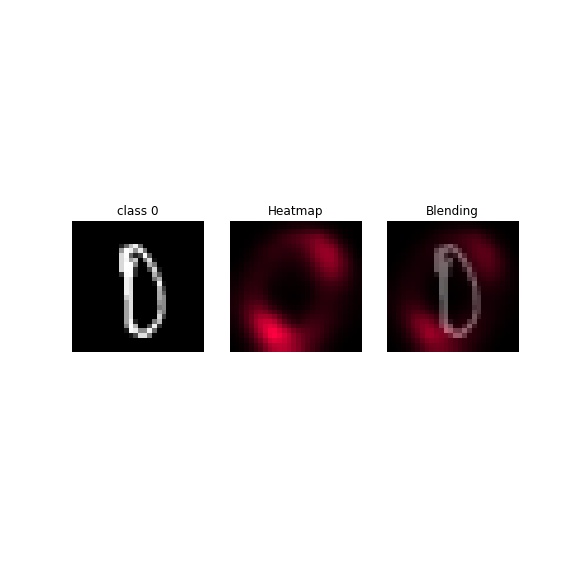}\\	
		\includegraphics[width=\linewidth, trim=70 210 50 200,clip]{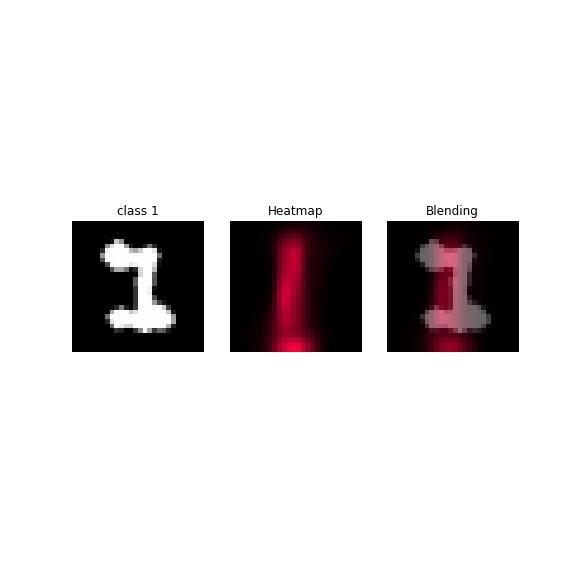}\\
		\includegraphics[width=\linewidth, trim=70 210 50 200,clip]{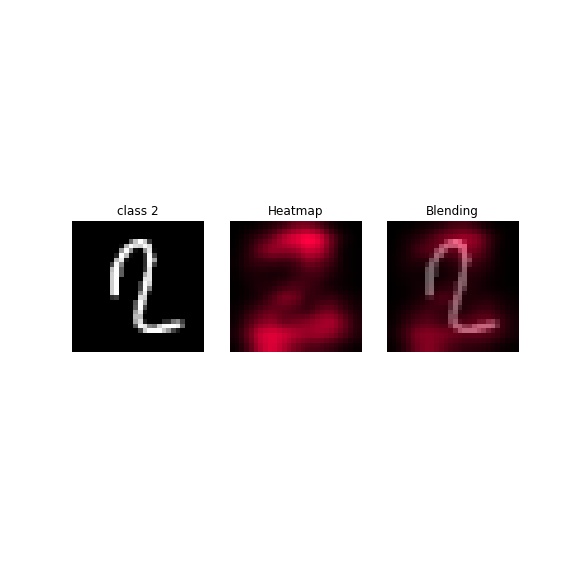}\\	
		\includegraphics[width=\linewidth, trim=70 210 50 200,clip]{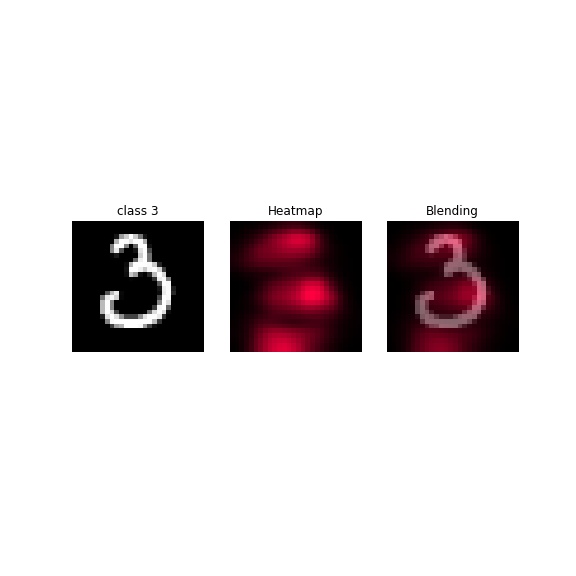}
		\caption{Low confidence samples}
		\label{fig:heatmap2}
	\end{subfigure}
	\caption{Model Attention Map}
	\label{fig:heatmap}
\end{figure}

To have an intuitive understanding on how a model behaves on different data styles, we use heatmaps \cite{heatmap} to present the model focus. In Figure \ref{fig:heatmap}, the first columns present images sampled from the training data according to their confidence scores. High confidence samples are shown in Figure \ref{fig:heatmap1}, and low confidence samples are shown in Figure \ref{fig:heatmap2}. The second columns in the figures are an average of 500 random heatmaps for each class which considered as the important area that a model focuses on. The two first columns are then blended and shown as the third columns. From the blending, it is clear that while the model heatmaps nicely fit high confidence samples, they poorly match the low confidence ones. This phenomenon can be interpreted as the model's attention is dominated by the most seen styles in the training data. In addition, we observe that low confidence scored samples have uncommon styles that barely appear in the training data as some examples are shown in Figure \ref{fig:confidence}. This consolidates that the model is not confident predicting these styles which is lacking of training representatives. For this reason, data diversity with limited representatives might reduce a machine learning model performance.

\section{\Methodname for Data Diversity}
\label{sec:method}
   
To alleviate the problem of deep learning performance reduction due to data diversity mentioned above, we propose a simple but efficient method namely \Methodname. The analysis in \ref{sec:analysis} suggested us to focus the blind spot of a model which is the uncertainty of the model. Since the model weights are dominated by the most data styles appearing in the training set, it might result in a poor prediction on other styles. We propose a simple method in which reconsidering these samples with another model to complement the previous model. We directly utilize the predictions from a strong model for isolating these samples. We then reuse them to train additional models to complement the previous strong model. This process can be conducted recursively resulting in a series of complementary models complementing each other. 

\begin{figure}[H]
	\centering
	\includegraphics[width=0.9\linewidth, trim=100 10 100 10,clip]{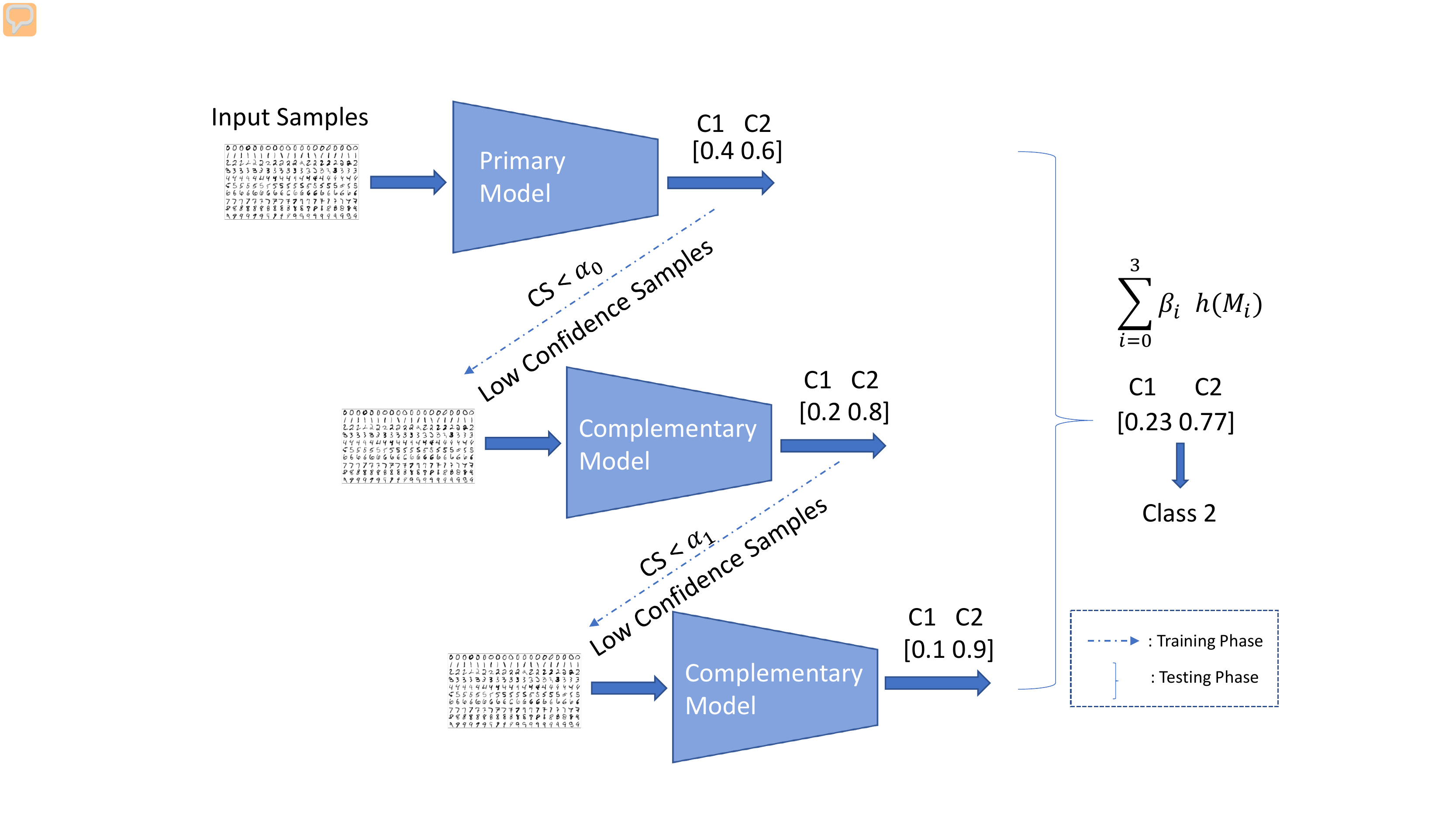}
	\caption{\Methodname }
	\label{fig:deepCE}
\end{figure}

The training process of \Methodname is described as follows. The training set is split to training and validation sets. First, we train a strong model named primary model. Then, the training data are predicted with the primary model resulting in a prediction and a corresponding confidence score for each sample. From the result, low confidence scored samples will be reused to train another model, namely complementary model. The term 'low confidence score' is determined by a threshold that could be tuned based on validation performance to achieve the best performance. In practice, the number of low confidence samples is relatively small, hence transfer learning is applied in this step. Primary model's weights are transferred to the complementary model, and the complementary model is trained using low confidence samples. This process can be repeated sequentially resulting in other children complementary models which complement their parents as shown in Figure \ref{fig:deepCE}. 

\begin{algorithm}[H]
	\caption{Algorithm for training \Methodname.}
	\begin{algorithmic}[1]
		\renewcommand{\algorithmicrequire}{\textbf{Input:}}
		\renewcommand{\algorithmicensure}{\textbf{Output:}}
		\REQUIRE Training dataset $X_0$. \\Parameter: Confidence Score Thresholds $\alpha_i$ \\ 
		\ENSURE  Primary and Complementary models
		\begin{itemize}
			\item \textit{Select a strong primary model $M_0$} 
			\item \textit{$i \leftarrow 1$}
		\end{itemize}
		\STATE  Train the primary model.\\		
		\FOR {$i$ in number of models}
		\STATE  Determine low confidence samples $X_{i}$
		\FOR {sample in $X_{i-1}$}			
		\STATE $cs \leftarrow \theta(M_{i-1}, sample)$
		\STATE if $cs  \leq \alpha_i$, append the sample to $X_i$
		\ENDFOR
		\STATE  Transfer the $(i-1)^{th}$ model's weights to the $i^{th}$ model
		\STATE  Train the $i^{th}$ complementary model with $X_i$ 	
		\ENDFOR
		\RETURN 
	\end{algorithmic}
	\label{alg:train}
\end{algorithm}

We formulate \Methodname along lines in Algorithm \ref{alg:train}. Let $M_0$ be the primary model and $M_1$ to $M_m$ be m sequential children models. $\alpha_0$ to $\alpha_n$ are the corresponding low confidence thresholds for determining samples reused in training children complementary models. Let $X_0$ be the full training data, and $X_i$ be the subset of $X$ corresponding to training data for the $i^{th}$ complementary model. $\theta(M_i,s)$ is the confidence score function of the $i^{th}$ model and input sample $s$. After selecting and training a strong primary model in line 1. We then sequentially train complementary models (lines 2-8). For the $i^{th}$ complementary model, we determine the training dataset from its parent model. The training dataset $X_{i-1}$ are predicted by the $(i-1)^{th}$ model and confidence scores are achieved (line 5). The samples that were predicted with low confidence will be added to training set $X_i$ as in line 6. Finally, we apply transfer learning to the $i^{th}$ complementary model, and train it with $X_i$.    

During the testing phase, all model outputs are assembled; the final label will be determined as the class has most significant probability-like output. Let $h(M,s)$ be the softmax output of model $M$ for sample $s$. The predicted class $ \mathcal{C} $ can be described in Equation \ref{eq:ensemble}. 
\begin{equation}
  \mathcal{C} = \argmax{\sum_{i=0}^{m}{\alpha_ih(M_i,s)} }
\label{eq:ensemble}
\end{equation}

A simple \Methodname example for a binary classification problem is visualized in Figure \ref{fig:deepCE}. The example uses three models, includes a primary model and two complementary models. The primary model is fully trained resulting model parameters and low confidence training samples with $CS < \alpha_0$. A complementary model is trained with the low confidence samples taken from primary model. Then, another complementary model is trained with low confidence samples taken from the above complementary model. After training all models, the final result in testing phase is an ensemble method of the three models weighted evenly with the $\beta$s' values of 0.33.

\section{Experiments}
\label{sec:experiments}

In this section, we apply \Methodname on classification tasks for both low resolution image datasets (i.e., MNIST and Fashion MNIST) and high resolution image datasets (i.e., Eardrum, Caltech and Chest-Xray). We selected state of the art models that we considered as the most efficient architectures for each dataset and enhance the model performance using our method. Specifically, we apply our method to a basic CNN for MNIST and Fashion MNIST dataset. Similarly, we apply \Methodname to several state-or-the-art deep learning models for Eardrum, Caltech and Chest-Xray dataset.

\begin{table*}[bp]
	\caption{Eardrum Classification Accuracy }
	\begin{threeparttable}
		\centering
		\begin{tabular}{|c|c|c|c|c|c|c|}
			\hline	
			& \textbf{Models} & \textbf{Resnet152V2}& \textbf{Xception} &  \shortstack{ \textbf{Inception}\\ \textbf{Resnet}} &\textbf{InceptionV3}  & \textbf{VGG16} \\
			\hline
			&Input Size        & 224x224 & 224x224 & 299x299 & 299x299 &  224x224 \\
			\hline \cline{2-7} 
			\multirot{2}{20}{Eardrum} & Acc. IM        & 83.13 & 81.04 & 81.88 & 81.67 & 78.70 \\ \cline{2-7}  	
			&Acc. CE & \textbf{84.40} & \textbf{81.80} & \textbf{81.90} & \textbf{82.30} & \textbf{79.50} \\
			\hline \cline{2-7}
			\multirot{2}{20}{Caltech} & Acc. IM       & 98.49 & 98.49 & 98.11 & 97.74 & 95.47 \\ \cline{2-7}  	
			&Acc. CE & \textbf{98.49} & \textbf{98.51} & \textbf{98.49} & \textbf{97.79} & \textbf{95.50} \\
			\hline \cline{2-7}
			\multirot{2}{20}{ChestXray} & Acc. IM       & 83.07 & 83.81 & 84.78 & 83.65 & 86.50 \\ \cline{2-7}  	
			&Acc. CE & \textbf{87.02} & \textbf{86.06} & \textbf{86.22} & \textbf{86.06} & \textbf{86.54} \\
			\hline
		\end{tabular}
		\begin{tablenotes}
			\item Acc. IM: Accuracy of Individual Model (\%)
			\item Acc. CE: Accuracy of Complementary Ensemble (\%)
		\end{tablenotes}
	\end{threeparttable}%
	\label{tab1}
\end{table*}

\subsection{ \Methodname on MNIST and Fashion MNIST datasets}

MNIST contains 60,000 gray images of 10 handwritten digits (0-9) with image size of 28\x28 pixels. The data is split into three parts, specifically training data (40,000 samples), validation data (10,000 samples) and testing data (10,000 samples). We build a convolutional network (CNN) following the MNIST example in Keras website which was mentioned in \ref{sec:analysis}. In this experiment, we use one complementary model which has the same architecture with the CNN mentioned above. Both $\beta$s are set to 0.5. We achieved an accuracy of 99.21\% and \textbf{99.41\%} for the primary model and the \Methodname results, respectively. We can see that there was an improvement after applying \Methodname.

Similar to MNIST, Fashion MNIST \cite{fashion} includes 60,000 28\x28 grayscale images of 10 clothing objects. We split the dataset the same way, and use the same CNN structure to MNIST dataset. Our objective becomes a 10-class classification. While the single CNN can achieve an accuracy of 92.81\%, applying \Methodname can improve it to \textbf{93.45\%}. 

\subsection{\Methodname on Eardrum, Caltech and Chest-Xray}
The Eardrum dataset contains 325 labeled images (512x512 pixels) collected from different resources, i.e., 282 published images \cite{eardrum_turkey} and 43 images of our collection. The data are enlarged using augmentation techniques by rotating and flipping images. The dataset was originally published in seven categories of diseases; in these experiments, we combine all the diseases as an abnormal condition. We then aim to tackle the problem of otitis media classification, specifically detecting a normal and abnormal eardrum from an individual sample.

The Caltech101 \cite{caltech101} contains 9,145 images of objects belonging to 101 categories. In our experiments, we use the first 10 categories which include 1,245 images. The size of each image is roughly 300\x200 pixels. We also perform a multiclass classification task on this dataset with the state of the art deep learning models.  

In addition, we apply our method on another high resolution dataset, Chest-Xray \cite{chestxray}. The dataset contains 5,863 Xray images and 2 categories (Pneumonia and Normal). All the images are in high resolution, roughly 1800\x1300 pixels. We then perform a binary classification task on this dataset to determine an individual's lung condition.     
 
Different to MNIST and Fashion MNIST dataasets, theses datasets include high resolution images. Thus, we are able to apply the most recent state-of-the-art models without hurting task performance. Specifically, we reproduce the architectures of InceptionV3 \cite{inception}, Xception \cite{xception}, InceptionResnetV2 \cite{inceptionresnet}, Resnet152V2 \cite{resnet}, VGG16 \cite{vgg}. To achieve a good performance, the images are then resized to the input size suggested for each model.

Table \ref{tab1} shows the accuracy results for the state-of-the-art models and our \Methodname on the Eardrum, Caltech101 and Chest-Xray datasets. It is clear that our \Methodname can improve the performance of all individual models despite having different architectures and input sizes . \Methodname with Resnet152V2 outperformed other networks on Eardrum and ChestXray, and Xception outperformed others over Caltech dataset.

\section{Conclusion and Future Work}    
\label{sec:conclusion}     
 
We provided an exploration into how data diversity affects CNN model uncertainty and an efficient technique to overcome its implications. Specifically, a diversity of data styles and lack of data representatives might lead to model uncertainty and misclassification. The experiment results show that our proposed method not only improves machine learning performance for large and rich dataset such as MNIST, it also improves machine learning performance for limited representative dataset, i.e., Eardrum and Chest-Xray, Caltech datasets. Our technique, however, has not considered noise in training data. The noise might be confused with style differences of data, and it might reduce our technique efficiency. In future work, we would like to investigate sophisticated techniques to reduce effects of noise to our technique.

\section*{Acknowledgment}

Effort sponsored in whole or in part by United States Special Operations Command (USSOCOM), under Partnership Intermediary Agreement No. H92222-15-3-0001-01. The U.S. Government is authorized to reproduce and distribute reprints for Government purposes notwithstanding any copyright notation thereon.  
{\footnote{ The views and conclusions contained herein are those of the authors and should not be interpreted as necessarily representing the official policies or endorsements, either expressed or implied, of the United States Special Operations Command.} }

\bibliographystyle{IEEEtran}
\bibliography{citation}

\end{document}